\documentclass[sigconf]{acmart}

\AtBeginDocument{%
  \providecommand\BibTeX{{%
    \normalfont B\kern-0.5em{\scshape i\kern-0.25em b}\kern-0.8em\TeX}}}

\setcopyright{acmcopyright}
\copyrightyear{2018}
\acmYear{2018}
\acmDOI{10.1145/1122445.1122456}

\acmConference[Woodstock '18]{Woodstock '18: ACM Symposium on Neural
  Gaze Detection}{June 03--05, 2018}{Woodstock, NY}
\acmBooktitle{Woodstock '18: ACM Symposium on Neural Gaze Detection,
  June 03--05, 2018, Woodstock, NY}
\acmPrice{15.00}
\acmISBN{978-1-4503-XXXX-X/18/06}

\acmYear{2020}
\acmConference[Proceedings of the CIKM 2020 Workshops]{}{October 19-20, 2020}{Galway, Ireland}
\acmBooktitle{Proceedings of the Workshop on the 29th ACM International Conference on Information and Knowledge Management (CIKM 2020), October 19-20, 2020, Galway, Ireland}

\usepackage{algorithm, algorithmicx, algpseudocode}
\usepackage{subfigure,bigstrut}

\begin{document}

\title{Fooling Object Detectors: Adversarial~Attacks~by~Half-Neighbor~Masks}

\author{Yanghao Zhang}
\authornote{Both authors contributed equally to this research. This work is done when Fu Wang was visiting the Trustworthy AI Lab at University of Exeter.}
\affiliation{%
  \institution{University of Exeter}
  \city{Exeter}
  \state{EX4 4QF}
  \country{UK}
}
\email{yanghao.zhang@exeter.ac.uk}

\author{Fu Wang}
\authornotemark[1]
\affiliation{%
  \institution{Guilin Univ. of Electronic Technology}
  \city{Guilin, Guangxi}
  \state{541004}
  \country{China}
}
\email{fuu.wanng@gmail.com}

\author{Wenjie Ruan}
\authornote{Corresponding author. This work is supported by Partnership Resource Fund (PRF) on Towards the Accountable and Explainable Learning-enabled Autonomous Robotic Systems from UK EPSRC project on Offshore Robotics for Certification of Assets (ORCA) [EP/R026173/1], and the UK Dstl project on Test Coverage Metrics for Artificial Intelligence.}
\affiliation{%
  \institution{University of Exeter}
  \city{Exeter}
  \state{EX4 4QF}
  \country{UK}
}
\email{w.ruan@exeter.ac.uk}

\renewcommand{\shortauthors}{Zhang and Wang, et al.}

\begin{abstract}
Although there are a great number of adversarial attacks on deep learning based classifiers, how to attack object detection systems has been rarely studied.
In this paper, we propose a Half-Neighbor Masked Projected Gradient Descent (HNM-PGD) based attack, which can generate strong perturbation to fool different kinds of detectors under strict constraints. 
We also applied the proposed HNM-PGD attack in the CIKM 2020 AnalytiCup Competition, which was ranked within the top 1\% on the leaderboard. We release the code at \url{https://github.com/YanghaoZYH/HNM-PGD}.
\end{abstract}




\maketitle

\section{Introduction}

Object detection is one of the most fundamental computer vision tasks,
which not only performs image classification~\cite{zhang2018gnn, zeng2018collaboratively} but also identifies the locations of the objects in an image.
Now object detection has been widely applied as an essential component in many applications that requires a high-level security, such as identity authentication~\cite{SharifBBR16}, autonomous driving~\cite{MullerBCFC06}, and intrusion detection \cite{LuSF17}. 
In recent years, we witness the significant progress has been made in object detection, especially by taking the advantage of deep learning models. 
However, deep learning based object detection systems are also demonstrated to be vulnerable to adversarial examples \cite{LuSF17}.
The adversarial example was first identified by~\citet{SzegedyZSBEGF13}, primarily on classification tasks, they showed that maliciously perturbed examples can fool a well-trained Deep Neural Network (DNN) to output wrong predictions.
After that, a great number of methods have been proposed to generate adversarial examples~\cite{huang2020survey,ZhangRFH20}, notably such as 
First Gradient Sign Method~\cite{GoodfellowSS14} and 
Projected Gradient Descent (PGD)~\cite{MadryMSTV18}.
At the same time, some studies show that DNN based object detection models are also facing the same threat~\cite{LuSF17,HuangKR+18,YanWLY20}.

In this paper, we introduce an adversarial attack framework, called HNM-PGD, which can fool different types of object detectors under two strict constraints concurrently.
Our method first identifies a mask that meets the constraints, and then generates an adversarial example by perturbing a specific area that is constrained by the mask. 
Adversarial examples generated in this way are guaranteed to satisfy the limitation in terms of the number of perturbed pixels and connectivity regions, while remaining a high efficiency.
One key novelty in HNM-PGD lies on that it enables an automatic process without handcraft operation, which provides a practical solution for many real-world applications. 
As a by-product of this attack strategy, we found that some perturbations contain clear semantic information, which are rarely identified by previous studies and provide some insights regarding the internal mechanisms of object detectors.

\begin{figure*}[ht]
    \centering
    \includegraphics[width=0.9\textwidth]{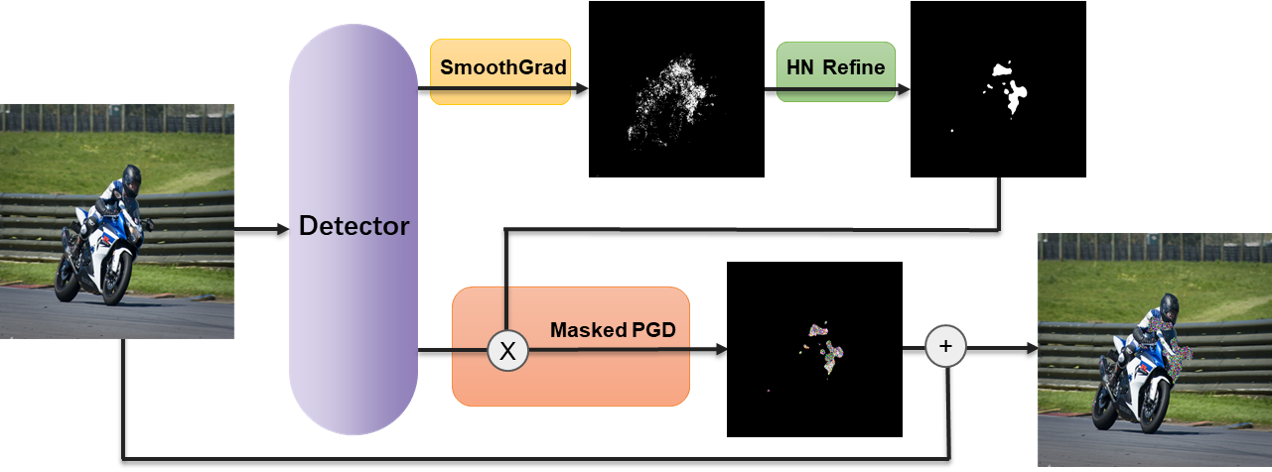}
    \caption{An illustration for the workflow of the proposed HNM-PGD.}
    \label{pipline}
\end{figure*}

\section{Background}
\subsection{Object Detection Models}
\label{background_obj}
Given an input example $x$, an object detector can described as $f(x)=z$, where $z$ represents the output vector of the detector.
Considering YOLOv4~\cite{bochkovskiy2020YOLOv4} and Faster RCNN~\cite{ren2015faster} as our target models, the information contained in $z$ is slightly different, and as an adversary under white-box setting, our goal is to make target models fail to detect the objects in the given examples.
Thus we focus on the target models' confidence about the existence of objects in $x$.
For each pre-defined box, YOLOv4 directly outputs its confidence $z^\text{conf} \colon \mathbb{R}$ about there is an object inside this box.
If $z^\text{conf}$ is above the given threshold, YOLOv4 model views this box as a potential object container, i.e. the area that may include objects.
Faster RCNN does not output $z^\text{conf}$, nevertheless, it introduces an extra background class and make predictions based on its classification result $z^\text{cls} \colon \mathbb{R}^{C+1}$, where $C$ is the number of classes.
Suppose $z^\text{cls}_i$ is the maximal item in $z^\text{cls}$, if $z^\text{cls}_i$ is greater than a given threshold and $i \neq C+1$, then the corresponding box will be viewed as the potential container.

\subsection{Constraints of Perturbation}
\label{background_cons}
In this paper, the restrictions of the adversary are 
\textit{i)} the number of perturbed pixels is not more than 2\% of the whole;
\textit{ii)} the number of 8-connectivity regions is not greater than 10.
Except for these two constraints, there are no limitations on the magnitude of the adversarial perturbation.
Because both constraints are related to the number of pixels, this belongs to the $\ell_0$ norm attack.

\subsection{Salience Map}
\label{background_salmap}
Salience map is a common tool to analyze and interpret DNN models' behaviors.
\citet{smilkov2017smoothgrad} proposed SmoothGrad method to generate stable salience maps.
Given a loss function $L$, the salience map is given by
\begin{equation}\large
S_x=\frac{1}{n} \sum_{i=1}^{n} \nabla_{x} L\left(f(x + \eta_i)\right),
\end{equation}
where $\eta_i$ are white noise vectors that sampled i.i.d. from a Gaussian distribution.

\section{Methodology}

\begin{algorithm}[t]
 \caption{Half-Neighbor Masked PGD (HNM-PGD)}
 \label{mask_pgd}
 \begin{algorithmic}[1] 
    \Require A given example $x$, number of random initialization $n$, control parameter $\phi$, HN kernel size $k$ and adjust step $s$, number of PGD iterations $P$, PGD step size $\alpha$
    \Ensure $\delta$
    \State$S_x=\frac{1}{n} \sum_{i=1}^{n} \nabla_{x} L\left(f(x + \eta_i)\right)$
    \Repeat 
    \State $z_\text{resp} = \operatorname{Mean}(S_x) + \phi \operatorname{Std}(S_x)$
    \State Initialize mask $M$ via $z_\text{resp}$
        \While{$k > 3$}
            \State $M = \text{HN}(M,k)$
            \State $M = \text{HN}(M,3)$
            \State $k = k - s$
        \EndWhile
    \State $\phi = \phi + 0.1$
   \Until{$M$ meets constraints}

    \State Random initialize $\delta$
    \State $\delta = \delta \times M$
    \Comment{$\times \colon$ element-wise product}
    \For{ $i = 1 \ldots P$}
         \State $\delta=\delta+\alpha \cdot \operatorname{sign}\left(\nabla_{\delta} L\left(f(x+\delta)\right)\right)$
         \State $\delta = \delta \times M$
        \State $\delta=\max (\min (\delta, \; 0-x), \; 1-x)$
    \EndFor
 \end{algorithmic}
\end{algorithm}

\begin{figure*}[!th]
        \centering
        \subfigure[ ]
        { \label{compimagenet}
        \includegraphics[width=0.3\textwidth]{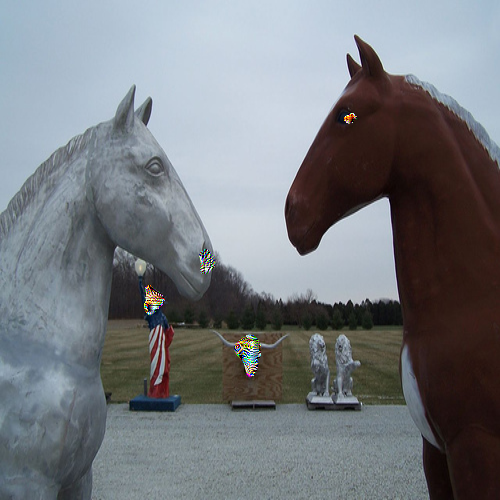}
        }
        \subfigure[ ]
        { \label{mse}
        \includegraphics[width=0.3\textwidth]{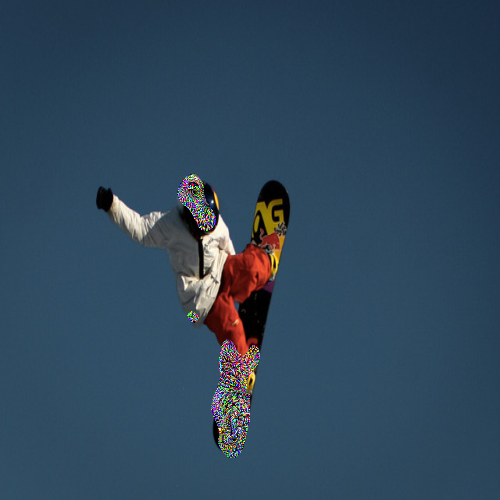}
        }
        \subfigure[ ]
        { \label{bar}
        \includegraphics[width=0.3\textwidth]{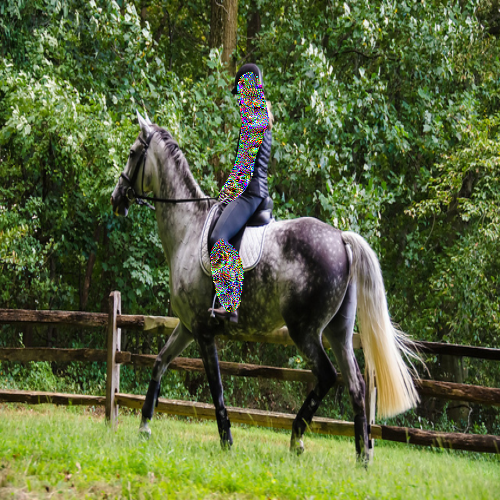}
        }
        \vspace{-3mm}
        \caption{Adversarial patches examples generated by the proposed HNM-PGD, the generated patches are mostly located in the semantic part of the object, such as the horse's eye, skateboard and human's body.} 
        \label{example}
        \vspace{3mm}
\end{figure*}


\subsection{Mask Finding}
We first propose a mask generation method to locate perturbation regions for any given examples.
Apparently, the size and shape of perturbation regions are critical to conduct a successful adversarial attack under the constraints.
Therefore, we use salience map to capture the model's response toward each pixel in $x$ at beginning. 
After compute an example's salience map, we initialize the mask via only keep pixels that the model's respond is larger than a threshold $z_\text{resp}$.
To automatically carry out this initialization, we borrow the idea of standard deviation and coverage from Gaussian distribution, and compute $z_\text{resp}$ via the mean and standard deviation of $S_x$, which can be described as
$
z_\text{resp} = \operatorname{Mean}(S_x) + \phi \operatorname{Std}(S_x),
$
where $\phi$ is a control parameter.

To meet the pixel constraints, we follow the spirit of K Nearest Neighbor algorithm to refine the mask.
Specifically, if half of a pixel's neighbors have been chosen by the current mask, then this pixel would also be chosen, otherwise it will be discarded.
We employ two convolution kernels whose parameters are all 1 to conduct this Half-Neighbor (HN) procedure. 
The first kernel aims to reduce the number of pixels in a mask, and its size is gradually changed during iterations. 
The second kernel is fixed to 3$\times$3, it can guarantee that there are no isolated pixels in the mask and reduce the number of connectivity regions (See lines 5--9 in Algorithm~\ref{mask_pgd}).
If the mask still does not meet the constraints, the algorithm will adjust $\phi$ accordingly and search again.

\subsection{Masked PGD Attack}
Once the perturbation regions are located, we generate adversarial examples via PGD iterations.
The basic idea here is summarized in Algorithm~\ref{mask_pgd}, where the selected regions are perturbed by a PGD adversary via conducting element-wise products between perturbation $\delta$ and mask $M$.
The workflow of the proposed defense is shown in Fig.~\ref{pipline}.

Due to the difference in the object detectors' output $z$, we need to consider YOLOv4 and Faster RCNN separately.
As we discussed in section~\ref{background_obj}, YOLOv4 directly outputs its confidence, so Binary Cross-Entropy (BCE) loss is a suitable option to conduct adversarial attack.
Suppose there are $m$ pre-defined boxes, and the maximal confidence is 1, BCE loss can be simplified as
\begin{equation}\large
\label{yolo_loss}
    L_{yolo}(z) = \sum_{i=1}^{m} \log z^\text{conf}_i,
\end{equation}
where $z$ is the output of a detector and $z^\text{conf}_i \in z$.

Different from YOLOv4, there are $C+1$ classes in Faster RCNN's classification result, including $C$ foreground objects and 1 background class.
To force the detector to classify an adversarial example into the background class, we conduct a targeted adversarial attack with a negative Cross Entropy (CE) loss, which can be written as
\begin{equation}\large
\label{frcnn_loss}
    L_{frcnn}(z) =z^\text{cls}_{C+1}-\log \Big(\sum_{j} \exp (z^\text{cls}_j)\Big),
\end{equation}
where $z^\text{cls}$ is the classification output of Faster RCNN detector, and $z^\text{cls} \in z$.
Note that we can attack YOLOv4 and Faster RCNN simultaneously by simply using HN masked PGD mixmize $L_{yolo} + L_{frcnn}$.

\section{Experiments}
To demonstrate our method, we select 100 images from MS COCO dataset as a toy dataset and conduct experiments for comparison on two white-box models, i.e. YOLOv4 and Faster~RCNN.

\subsection{Implementation Details}
\textbf{YOLOv4~}
Regarding the provided model YOLOv4, the input size is set to 608$\times$608 while the original image has 500$\times$500, to allow differential, we employ the function $\textbf{torch.nn.Upsample}$ with bilinear interpolation to resolve the resize problem. 
Due to the approximation computation of $\textbf{torch.nn.Upsample}$, we need to allow more boxes to be detected to stabilize the adversarial perturbation. 
As YOLOv4 method only outputs the foreground objects with $z^\text{conf} > 0.5$, we adjust the confidence threshold from $0.5$ to $0.3$ during attack.\\ 
\textbf{Faster RCNN~}
Similar to the configuration of YOLOv4, we resize the input to 800$\times$800 with bilinear interpolation.
As the permitted threshold for Faster RCNN is 0.3, which is relatively lower than YOLOv4. 
In practice, we assign a smaller threshold 0.1 when calculating the loss to enable more boxes to appear.\\
\textbf{PGD Settings~}
In this paper, the HNM-PGD is carried out with 40 steps, and the step size is $16/255$.
\begin{figure}[!t]
    \centering
    \includegraphics[width=0.85\columnwidth]{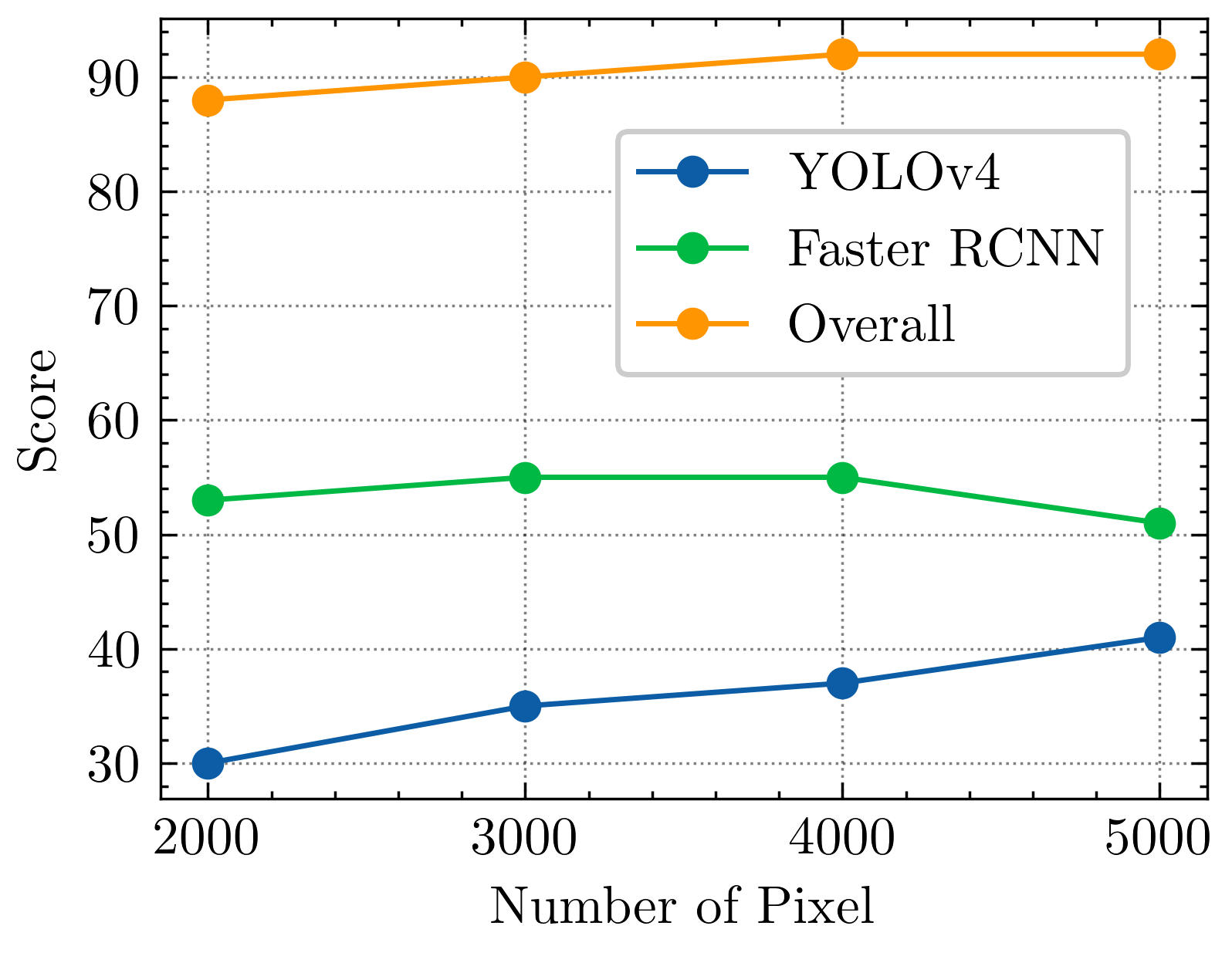}
    \caption{Performance on the toy dataset with the increasing amount of pixel.}
    \label{score}
\end{figure}

\subsection{Experimental Results}

In this part, we employ the formula in the description of AnalytiCup to calculate the score, which provides a criteria to evaluate the performance of the proposed method.
Our code is available on Github\footnote{\url{https://github.com/YanghaoZYH/HNM-PGD}}.

We produce 100 adversarial examples on the white-box models with two loss together: $L_{yolo} + L_{frcnn}$. 
Figure~\ref{example} gives several successful examples for the targeted models.
We can observe that the proposed method does locate the object correctly, and the added patches normally target on their sensitive parts.
Figure~\ref{score} illustrates the overall score with the increasing upper bound of the number of pixel among 100 selected images, and the scores achieved by YOLOv4 and Faster RCNN, respectively.
We find that with the increase of the quantity of pixel, Faster RCNN performs better, while this is not the case for YOLOv4.
This is because the provided white-box Faster RCNN uses a low tolerate threshold, where sufficient pixel is needed for successful attack. 
In terms of YOLOv4, the performance fluctuates at about 53. 
Therefore, there is a trade-off when choosing the amount of the pixel.

We apply the same strategy and perform the adversarial attack with more steps (800) and smaller step size ($4/255$) for all 1000 images under the different quantity of pixel, then we pick the best result obtained on the white-box models as our solution. 
In the final stage of the AnalytiCup competition, we had also tried to improve the generalization of the attacking approach on the unseen model, i.e. black-box. 
In detail, we add some transformations (like flipping/cropping) to the input image, which is expected to not overfit the known white-box models too much.
Our final score is 2414.87, ranked 17 in the competition.

\section{Beyond Competition}

This competition leads to a few interesting research directions.
Intuitively, due to the $\ell_0$ norm constraint, both location and shape of the perturbation are critical to the attacking performance.
We have reviewed other top contestants' solutions and found that linear adversarial patches have a higher impact on the target model's output and use less number of pixels than blocky ones, while location is less important.
This seems because blocky perturbation can only influence a relatively small range of a convolution kernel's output, while linear perturbation can cross a wider area.
To verify such conjecture, we wish to adopt verification technologies on neural networks~\cite{WuWRHK20,RuanWSHK19,RuanHK18} into the object detectors and quantify the worst-case scenario of adversarial patches on object detectors.
Besides, on the top of our HNM-PGD, we can also expand evaluation of existing adversarial attacks and defenses, such as~\cite{ZhangRFH20,WangHLZ20}, onto object detection tasks. 

\section{Conclusion}

In conclusion, we propose a PGD-based approach to attack object detectors using Half-Neighbor masks.
In the proposed HNM-PGD, the automatic pipeline allows it to craft adversarial examples/patches automatically under $\ell_0$ constraint, which can be applied in many applications, even physical-world attacks. 
On the other hand, this end-to-end attack framework also benefits further studies on defending object detectors against adversarial attacks and verifying their robustness.

\section*{Acknowledgments}
The work was partially supported by
the Guangxi Science and Technology Plan Project under Grant AD18281065,
the Guangxi Key Laboratory of Cryptography and Information Security under Grant GCIS201817.
Fu Wang is supported by the study abroad program for graduate student of Guilin University of Electronic Technology under Grant GDYX2019025, and the Innovation Project of GUET Graduate Education under Grant 2020YCXS042.

\bibliographystyle{ACM-Reference-Format}
\bibliography{reference}
\end{document}